\documentclass[11pt,a4paper]{article}
\usepackage[hyperref]{eacl2021}
\usepackage{times}
\usepackage{latexsym}

\usepackage[normalem]{ulem}

\usepackage{enumitem}
\usepackage{dirtytalk}
\usepackage{colortbl}
\usepackage{xstring}

\usepackage{textcomp}
\usepackage{placeins}
\usepackage{subfigure}
\usepackage{multirow}
\usepackage{babel,blindtext}
\usepackage{balance}

\usepackage[justification=centering]{caption}   
\usepackage{graphicx}
\usepackage{epstopdf} 
\usepackage{float}  
\usepackage[export]{adjustbox}
\usepackage{rotating}
\usepackage{booktabs,threeparttable}
\usepackage{parskip}

\usepackage{placeins} 
\usepackage{xcolor}
\usepackage{graphicx,wrapfig,lipsum}
{}

\definecolor{darkgreen}{HTML}{036400}  
\definecolor{darkred}{HTML}{680100} 
\definecolor{OliveGreen}{cmyk}{0.64,0,0.95,0.40}

\usepackage{microtype}

\aclfinalcopy 
\def\aclpaperid{927} 

\title{A Study of Automatic Metrics for the Evaluation of Natural Language Explanations}

\author{Miruna-Adriana Clinciu \\
  Edinburgh Centre for Robotics \\
Heriot-Watt University \\ University of Edinburgh \\

  \texttt{mc191@hw.ac.uk} \\\And
  
  Arash Eshghi \\
  Heriot-Watt University \\ Edinburgh, United Kingdom \\
  \texttt{a.eshghi@hw.ac.uk} \\\And
  
  Helen Hastie \\
  Heriot-Watt University \\ Edinburgh, United Kingdom \\
  \texttt{h.hastie@hw.ac.uk} \\}
  
\date{}

\begin{document}
\maketitle
\begin{abstract}
As transparency becomes key for robotics and AI, it will be necessary to evaluate the methods through which transparency is provided, including automatically generated natural language (NL) explanations. Here, we explore parallels between the generation of such explanations and the much-studied field of evaluation of Natural Language Generation (NLG). Specifically, we investigate which of the NLG evaluation measures map well to explanations. We present the ExBAN corpus: a crowd-sourced corpus of NL explanations for Bayesian Networks. We run correlations comparing human subjective ratings with NLG automatic measures. We find that embedding-based automatic NLG evaluation methods, such as BERTScore and BLEURT, have a higher correlation with human ratings, compared to word-overlap metrics, such as BLEU and ROUGE. This work has implications for Explainable AI and transparent robotic and autonomous systems. 
\end{abstract}

\section{Introduction}
The machine learning models and algorithms underlying today's AI and robotic systems are increasingly complex with their internal operations and decision-making processes ever more opaque. This opacity is not just an issue for the end-user, but also the creators and analysts of these systems. As we move towards building safer and more ethical systems, this lack of transparency needs to be addressed. One key trait of a transparent system is its ability to be able to \textit{explain} its deductions and articulate the reasons for its actions in Natural Language (NL). As the area of Explainable AI (XAI) grows and is mandated (cf. the EU General Data Protection Regulation\textquotesingle s \say{right to explanation} \citep{gdprart22} and standardisation (cf. IEEE forthcoming standard on Transparency (P7001)), it has become ever more important to be able to evaluate the quality of the NL explanations themselves, as well as the AI algorithms they explain. Furthermore, the importance of evaluating explanations has been emphasised by researchers within the social cognitive sciences \citep{Leake2014, Zemla2017, Doshi-Velez2017}. To date, explanations have mostly been evaluated by collecting human judgements, which is both time-consuming and costly. Here, we view generating explanations as a special case of Natural Language Generation (NLG), and so we explore mapping existing automatic evaluation methods for NLG onto explanations. We study whether general, domain-independent evaluation metrics within NLG are sensitive enough to capture the peculiarities inherent in NL explanations \cite{Kumar2020}, such as causality; or whether NL explanations constitute a sui-generis category, thus requiring their own automatic evaluation methods and criteria.

In this paper, we firstly present the ExBAN dataset: a corpus of NL explanations generated by crowd-sourced participants presented with the task of explaining simple Bayesian Network (BN) graphical representations. These explanations were subsequently
rated for \emph{Clarity} and \emph{Informativeness}, two subjective ratings previously used for NLG evaluations \cite{Gatt2018, howcroft-etal-2020-twenty}. The motivation behind using BN is that they are reasonably easy to interpret, are frequently used for the detection of anomalies in the data \cite{Tashman2020, Saqaeeyan2020, Metelli2019, Mascaro2014}, and have been used to approximate deep learning methods \cite{Riquelme2018, Gal2016}, which we could, in turn, explain in Natural Language.

Secondly, we explore a wide range of automatic measures commonly used for evaluating NLG to understand if they capture the human-assessed quality of the corpus explanations.  We then go on to discuss their strengths and weaknesses through quantitative and qualitative analysis. 

Our contributions are thus as follows: (1) a new corpus of natural language explanations generated by humans, who are asked to interpret Bayesian Network graphical representations, accompanied by subjective quality ratings of these explanations. This corpus can be used in various application areas including Explainable AI, general Artificial Intelligence, linguistics and NLP; (2) a study of methods for evaluating explanations through automatic measures that reflect human judgements;
and (3) qualitative discussion into these metrics' sensitivity by examining specific explanations varying on the Informativeness/Clarity scales. 

\section{Related Work}

Explanations are a core component of human interaction \cite{Scalise2017, krening2017, Madumal2019}. In the context of Machine Learning (ML), explanations should articulate the decision-making process of an ML model explicitly, in a language familiar to people as communicators \cite{DeGraaf2017,DBLP:journals/corr/Miller17a}. According to \newcite{Plumb2018}, three of the most common types of explanation are: (1) \emph{global explanations}, which describe the overall behaviour of the entire model \cite{aix360-sept-2019}; (2) \emph{local explanations}, commonly taking the form of counterfactuals \cite{Sokol2019} that describe why particular events happened (known also as \say{everyday explanations}); and (3) \emph{example-based explanations} that present examples from the training set to explain algorithmic behaviour \cite{Cai2019}. 

Recently, various explanation systems provide different types of explanations for AI systems: the LIME method visually explains how sampling and local model training works by using local interpretable model-agnostic explanations \cite{Ribeiro2016}; MAPLE can provide feedback for all three types of explanations: example-based, local and global explanations \cite{Plumb2018}; CLEAR explains a single prediction by using local explanations that include statements of key counterfactual cases \cite{DBLP:journals/corr/abs-1908-03020}. 
Whilst these techniques and tools gain some ground in explaining deep machine learning, the explanations they provide are not necessarily aimed at the (non-expert) end-user and so are not always intuitive. 

NLG has traditionally been broken down into \say{what} to say (content selection) and \say{how} to say it (surface realisation) and can draw parallels with Natural Language explanations. In particular, it is important to gauge how much content or how many reasons to present to the user, to inform them fully without overloading them. For example, prior work has shown that people prefer shorter explanations that offer only sufficient detail to be considered useful \cite{Harbers2009,Yuan:2011:MRE:2208436.2208445}. 

According to \newcite{Miller2017}, how explainers generate and select explanations depends on so-called pragmatic influences of causes, and they found that people seem to prefer simpler and more
general explanations. Similarly, \newcite{Lombrozo2007} notes that simplicity and generality might be the key to evaluating explanations. This was partly the case described in \cite{GarciaetalINLG18}, but here the users were experts and preferred to be given all possible reasons but as precise and brief as possible.  It is clear from these prior works that explanations have to be evaluated in the context of the scenario, prior knowledge and preferences of the explainee, and the intent and goals of the explainer. These could be, for example, establishing trust \citep{Miller2017}, agreement, satisfaction, or acceptance of the explanation and the system \citep{Gregor1999}. 

Somewhat analogous to auto-generated explanations are the fields of summarisation of text \cite{Tourigny1998, Deutch2015} and Question-Answering \cite{Dali2009, Xu2017, lamm2020qed}. This is because they provide users (expert and lay users) with various forms of summaries (visual or textual) and answers containing explanations to enable them to have a better understanding of content.

Summarisation methods and sentence compression techniques can help to build comprehensive explanations \cite{Winatmoko2013}. 
With regards to evaluating these summarisation methods, \newcite{Xu2020} proposed an evaluation metric that weighted the facts present in the source document according to the facts selected by a human-written (natural language) summary, by using contextual embeddings. This evaluation of text accuracy is indeed related to explanations because any explanation must contain enough statements to support decision-making and understanding. These statements should be accurate and true.

The growing interest in the AI community to investigate the potential of NL explanations for bridging the gap between AI and HCI has resulted in an increasing number of NL explanations datasets. The ELI5 dataset\footnote{\url{https://facebookresearch.github.io/ELI5/}} \cite{fan2019eli5} is composed of explanations represented as multi-sentence answers for diverse questions where users are encouraged to provide answers, which are comprehensible for a five-year-old. WorldTree V2\footnote{\url{http://www.cognitiveai.org/explanationbank}} \cite{Jansen2019} is a corpus of Science-Domain that contains explanation graphs for elementary science questions, where explanations represent interconnected sets of facts. CoS-E\footnote{\url{ https://github.com/salesforce/cos-e}} is a dataset of human explanations for commonsense reasoning in the form of natural language sequences and highlighted annotations \cite{rajani2019explain}. Multimodal Explanations Datasets (VQA-X and ACT-X) contain textual and visual explanations from human annotators \cite{Park2018}. e-SNLI\footnote{\url{https://github.com/OanaMariaCamburu/e-SNLI}} is a corpus of explanations built on the question: \say{Why is a pair of sentences in a relation of entailment, neutrality, or contradiction?} \cite{Camburu2018}. Finally, the SNLI corpus\footnote{\url{https://nlp.stanford.edu/projects/snli/}} is a large annotated corpus for learning natural language inference \cite{Bowman2015}, considered one of the first corpora of NL explanations.

In this paper, we present a new corpus for NL explanations. The ExBAN corpus presented here provides a valuable addition to this set of corpora as it is aimed at explaining structured graphical models (in particular Bayesian Networks), that are closely linked to ML methods.

\FloatBarrier
\begin{figure*}[h!]
\centering
\includegraphics[width=\textwidth]{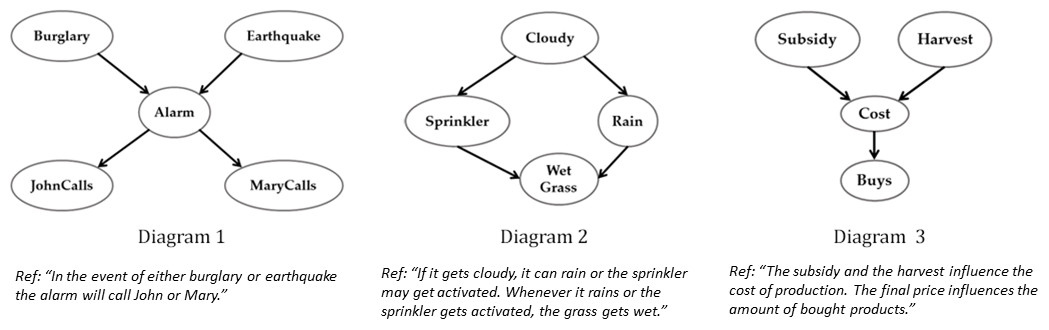}
\caption{Annotated diagrams with assigned explanation references, where {\bf Diagram 1} represents a typical Bayesian Network, {\bf Diagram 2} represents a multiply-connected network and {\bf Diagram 3} represents a simple network with both discrete variables (Subsidy and Buys) and continuous variables (Harvest and Cost). Beneath each diagram, the gold standard references are provided.}
\label{fig:diagrams}
\end{figure*}

\section{ExBAN Corpus}
\label{sec:corpus}
The {ExBAN} Corpus (\textbf{Ex}planations for \textbf{BA}yesian \textbf{N}etworks)\footnote{ The data is openly released at \url{ https://github.com/MirunaClinciu/ExBAN}} consists of NL Explanations collected in a two step process: (1) NL explanations were produced by human subjects; (2) in a separate study, these explanations were evaluated in terms of Informativeness and Clarity.

For Step 1, each subject was shown graphical representations of three Bayesian Networks (BN), in random order.  They were then asked to produce text to describe how they interpreted the BN. The three BN used in the data collection are presented in Figure \ref{fig:diagrams}and represent well-known BN examples, extracted from \newcite{russell2019human}.  For Step 2 in a separate experiment, approximately 80 of these generated explanations were presented to a different set of subjects in random order, along with a scenario description and the graphical model image. Subjects were asked to rate them in terms of Informativeness and Clarity. The worded scenario descriptions were not given to subjects in the first stage, so as not to prime them when generating explanations. 

\subsection{Step 1: Natural Language Explanations Corpus}
{\bf Survey Instrument.} A pilot was performed to test options and ensure the completion time, leading to the final survey instrument. The survey was divided into five sections: 1) consent form;
2) closed-ended questions related to English proficiency, computing and AI experience: \say{How much computing experience do you have?}, \say{What is your English Proficiency Level?}, \say{How much experience do you have in the field of Artificial Intelligence?}; 3) attention-check question, where participants received an image of a graphical model, and they had to select the correct answer(s) for the given image; and 4) respondents were asked to explain the three graphical models, in their own words. All respondents received the graphical model survey questions in randomised order. 
The appropriate ethical procedures were followed in accordance with ethical standards, and ethical approval was obtained. 

{\bf Participants.} 85 participants were recruited via social media. English proficiency level, computing experience and AI experience were rated on a numerical scale, from 1 to 7 (1 = beginner, 7 = advanced).  The majority of participants ($n = 83$) rated their level of English proficiency with values higher than 5, with over half of the participants rating their level as 7.  Just 12\% ($n = 10$) participants rated their computing experience scores with a value lower than 5 and 82\% ($n = 70$) of participants had a high level of computing experience. Subjects had mixed experience with AI with over half (54\%) having some experience ($n = 46$), but 46\% of them had limited AI experience ($n = 39$).

{\bf Collected NL explanations.} Quality control of the collected data included a cleaning step where participants' responses were hand-checked and removed if the participants did not attempt to complete the tasks. Explanations that contained misspellings and missing punctuation were corrected manually (both the raw data and cleaned data are available). The number of explanations for each diagram, after the data cleaning step are as follows: Diagram 1: 84 explanations, 1788 words; Diagram 2: 83 explanations, 1987 words; and Diagram 3: 83 explanations, 1400 words.

\subsection{Step 2: Human Evaluation for Quality}
{\bf Survey Instrument.} To investigate the quality of the explanations collected in Step 1, we performed a human evaluation of the generated explanations. A pilot survey was performed to test and refine options, where respondents ($n = 45$) were recruited from Amazon Mechanical Turk and were compensated monetarily.

Each participant was given three tasks, each corresponding to the BN presented in Figure \ref{fig:diagrams} with the order randomised. Along with the BN image, a simple description story was provided in order to give the subject a better understanding of the context as well as instructions on how to approach these tasks. Here, we give the scenario for Diagram 1 to illustrate this: \say{John and Mary bought their dream home. To keep their home safe, they installed a Burglary/Earthquake Alarm. Also, they received an instruction manual where they found the following diagram: They are not sure if they correctly understood the diagram. On the following pages are some worded explanations. We need your help to evaluate them!}

For every BN image, the participants were asked to evaluate 5 explanations in terms of: {\bf Informativeness} (Q: \say{How relevant the information of an explanation is}; Likert scale, where 1 = Not Informative and 7 = Very Informative); and {\bf Clarity} (Q: \say{How clear the meaning of an explanation is}; Likert scale, where 1 = Unclear and 7 = Very Clear).

{\bf Participants.} The final data collection survey was advertised on social media as \say{a 10-minute survey, where participants were asked to provide feedback about how understandable the explanations of the three graphs are}. Demographic information was collected (age range and gender). A total of 96 participants answered the survey. As screening criteria, participants had to complete all survey questions. Post validation, we had a sample of 56 participants consisting of 42 male participants (75\%), 11 female participants (19.6\%) and 2 non-binary gender participants (3.6\%). Gender imbalance might be due to \say{differences in female and male values operating in an online environment} \citep{Smith2008}. Half of the participants ($n = 28$) are aged between 23-29 years old, followed by 30\%  of participants aged between 18-22 ($n = 17$), 20\%  aged 40–49 ($n = 11$), 18\%  aged 30–39 ($n = 10$). Previous studies have identified a high degree of inconsistency in human judgements of natural language \citep{Novikova, Dethlefs2014}; each participant can have a different perception of the interpretation of these metrics, even if a definition of these metrics is provided. Indeed, we found that in our data,  explanation ratings can vary significantly, with an explanation rated highly by one person for Clarity, but viewed as very unclear by another annotator. This was the case for both Clarity and Informativeness. 
 
We aim to create a reliable database of varying quality of NL explanations, i.e. where the quality of explanations is generally uncontested by the majority. Therefore, subjective ratings were post-processed. For each explanation, we collected a minimum of 3 judgments. Explanations received ratings from 1 to 7; we classified bad explanations as those with low ratings (ratings $<$5) and good explanations, as those with higher ratings (ratings $\geq$5).
For any one explanation, if the difference between the number of good and bad ratings is $\leq$1, then that explanation is considered hard to judge and difficult to reach a consensus on and thus removed. After this pre-processing step, the corpus contained ratings for 54 explanations for Diagram 1, 34 explanations for Diagram 2, and 54 explanations for Diagram 3.

To verify the agreement between different raters, we used Krippendorff's Alpha, a measure of inter-rater reliability \cite{Krippendorff1980}. We computed Krippendorff's Alpha coefficient using the Python package \texttt{krippendorff} (version 0.3.2). After the pre-processing step, the agreement between subjects increased, see Table \ref{table:alpha} for the post-processing Alpha values for each of the Bayes Nets. Alpha values between .21 to .40 indicate fair agreement and values between .41 to .60 indicate moderate agreement \cite{Hallgren2012}. Here, we can see that explanations for Diagram 2 were particularly contentious, but overall the numbers reflect fair to moderate agreement. 
\begin{table}[!htb]
\centering
{
\scalebox{0.75}{
\begin{tabular}{lllll}
& \multicolumn{1}{c}{Diagram 1} & \multicolumn{1}{c}{Diagram 2} & \multicolumn{1}{c}{Diagram 3} & {All Diagrams} 
\\ \hline
Inform.
& 0.514
& 0.202
& 0.420
& 0.377
\\
Clarity
& 0.440
& 0.182
& 0.361
& 0.319
\end{tabular}}
\caption{
Inter-annotator agreement measured by Krippendorff’s Alpha}
\label{table:alpha}}
\end{table}

\section{NLG Evaluation Metrics}
Here, we describe the reasoning behind our choice of subjective measures that attempt to capture both the content and its correctness (Informativeness) and quality of expression (Clarity). We also describe objective measures commonly used for automatic evaluation of NLG, and which we will extract from the ExBAN corpus. 
\subsection{Subjective NLG Evaluation Metrics}

Human evaluation is considered a primary evaluation criterion for NLG systems \cite{Gatt2018, Mellish1998, gkatzia-mahamood-2015-snapshot, Hastie2014}. Through Explainable AI, we want to achieve Clarity and understanding in communicating the process of AI systems. Therefore, explanations should be clear and easily understood by users. Traditional human evaluation metrics are clearly needed for increasing transparency, avoiding confusion and misunderstanding.

{\bf Informativeness.} As defined in the field of NLG, Informativeness targets relevance or correctness of an NLG output relative to an input \cite{Dusek2020}. According to the literature, Informativeness can provide \say{timely, relevant and useful information} \cite{Novikova} and \say{make information immediately accessible} \cite{Maxwell2017}. Sometimes, Informativeness is linked with accuracy, or adequacy \cite{Novikova}. As mentioned previously, explanations contain statements with some prior knowledge that must be accurate and true \cite{Goodrich2019, Xu2020}. 

{\bf Clarity.} An explanation should be clear to achieve effective communication. In the NLG field, Clarity implies that text is easily understood \cite{Belz2009kow, VanDerLee2017} and that the reader is familiar with basic information introduced in the text \cite{Lampouras2013}. In addition, Clarity can also help expose the truthfulness and correctness of textual data \cite{Mahapatra2016}.

\subsection{Automatic Evaluation Metrics}
This section describes a number of automatic metrics commonly used in NLG evaluation and selected for this study. These fall into two categories: 1) word-overlap metrics, e.g. BLEU, METEOR and ROUGE \cite{novikova-etal-2017-need}; and 2) embedding-based metrics, e.g. BERTScore and BLEURT \cite{sellam-etal-2020-bleurt}. Each of these metrics is compared to one or more \say{Gold Standard} text as inspired by the Machine Translation community and adopted for evaluating document summarisation and NLG \citep{Belz2006}. The gold standard is normally a piece of natural language text, annotated by humans as correct, i.e. a solution for a given task. Automatic evaluation is based on this gold standard, 
by verifying potential similarity \citep{Kovar2016}. However, the selection of gold standards involves subjectivity and specificity \citep{Kovar2016}, and this is part of the reason that automatic metrics have received some criticism \cite{Hardcastle2008}. 

{\bf BLEU} (B) \citep{Papineni2001} is widely used in the field of NLG and compares n-grams of a candidate text (e.g. that generated by an algorithm) with the n-grams of a reference text. The number of matches defines the goodness of the candidate text. {\bf SacreBLEU} (SB) \citep{post-2018-call} is a new version of BLEU that calculates scores on the detokenized text. {\bf METEOR} (M) was created to try to address BLEU\textquotesingle s weaknesses \citep{lavie-agarwal-2007-meteor}. METEOR evaluates text by computing a score based on explicit word-to-word matches between a candidate and a reference. When using multiple references, the candidate text is scored against each reference, and the best score is reported. {\bf ROUGE} (R) \citep{Lin1971} evaluates n-gram overlap of the generated text (candidate) with a reference. {\bf ROUGE-L} (RL) (Longest Common Subsequence) computes the longest common subsequence (LCS) between a pair of sentences. 

{\bf BERTScore} (BS) \citep{bert-score} is a token-level matching metric with pre-trained contextual embeddings using BERT \citep{Devlin2019} that matches words in candidate and reference sentences using cosine similarity. {\bf BLEURT} (BRT) \citep{sellam-etal-2020-bleurt} is a text generation metric also based on BERT, pre-trained on synthetic data; it uses random perturbations of Wikipedia sentences augmented with a diverse set of lexical and semantic-level supervision signals. BLEURT uses a collection of metrics and models from prior work, including BLEU and ROUGE. Evaluation based on the meanings of words using embeddings (BERTScore, BLEURT) might capture some relevant features of explanations, as
word representations are dynamically informed by the words around them \cite{berttutorial}).

\section{Correlation Study of Automatic Metrics}
As noted in the introduction, it remains an open question as to what degree the automatic metrics for NLG reviewed above can capture the quality of NL explanations \cite{Clinciu2019}.
Thus, we ran a correlation analysis to investigate the degree to which each of the automatic metrics correlates with human judgements using the ExBAN corpus, and which aspects of human evaluation (Clarity/Informativeness), such automatic measures can capture.   With regards to the choice of gold standard text, we picked explanations that received the maximum score in the human evaluation, in both Clarity and Informativeness. Gold standard explanations of each diagram are presented in Figure \ref{fig:diagrams}. 

\subsection{Results}
\begin{figure}
\centering
\includegraphics[width=0.5\textwidth]{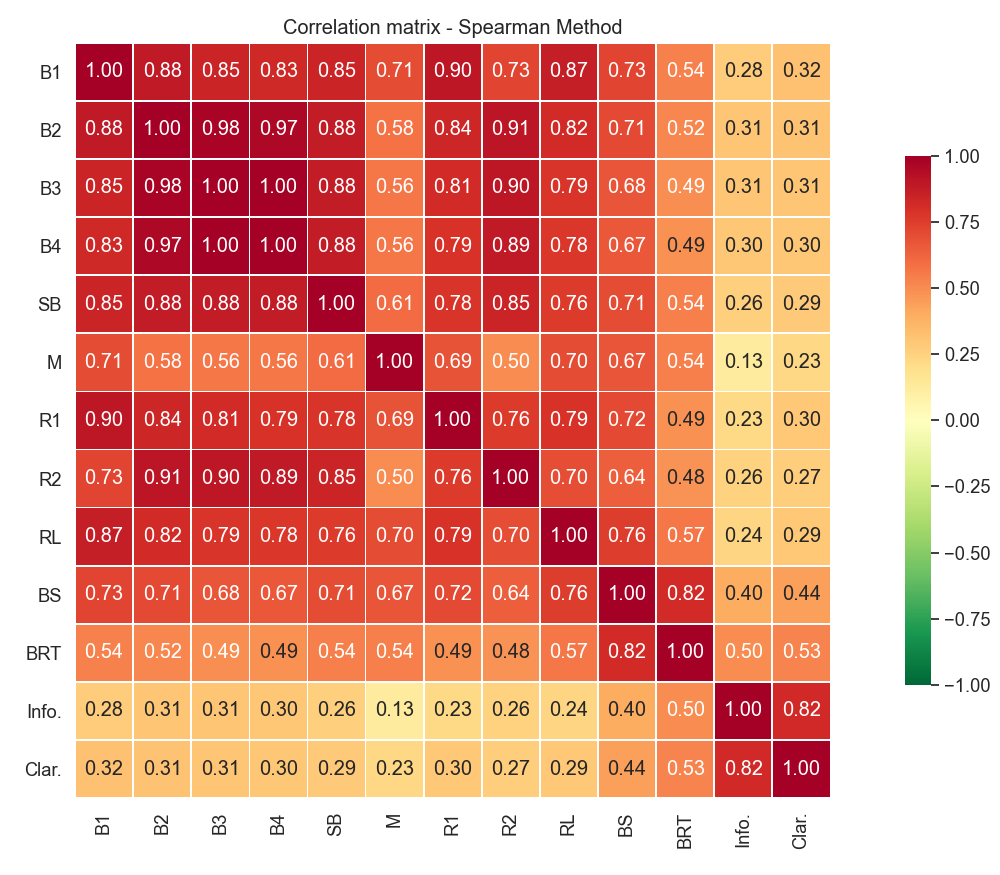}
    \caption{Heatmap of Spearman rank correlation between automatic evaluation metrics and human evaluation metrics (Informativeness and Clarity)}\label{fig:all_corr}
\end{figure}

The correlations between automatic metrics and human ratings were computed using the Spearman correlation coefficient. For each explanation, we calculated the median of all the ratings given (median was calculated because the data is ordinal, non-parametric rating data, as is also reported in
\newcite{Braun2018, novikova-etal-2017-need}).  These medians were then correlated with the automatic metric scores in Tables \ref{table:Informativenesscorr} and \ref{table:Claritycorr} and Figure \ref{fig:all_corr}. A summary of the results of the correlation analysis include the following:
\vspace{-0.5pc}
\begin{enumerate}[nosep]
    \item Word-overlap metrics such as BLEU~(n = 1,2,3,4), METEOR and ROUGE (n = 1,2) presented low correlation with human ratings.
    \item BERTScore and BLEURT outperformed other metrics and produced higher correlation with human ratings than other metrics on all diagrams. BERTScore values range between [0.23, 0.43] and for BLEURT values range between [0.26, 0.53]. 
    \item  Human ratings for Informativeness and Clarity are highly correlated with each other, as observed in Figure \ref{fig:all_corr} ($r = 0.82$).
\end{enumerate}

\FloatBarrier

\begin{table}[!htb]
\centering
{
\scalebox{0.65}{
\begin{tabular}{lllll}

Metric    & \multicolumn{1}{c}{\begin{tabular}[c]{@{}c@{}}Diagram 1\end{tabular}} & \multicolumn{1}{c}{\begin{tabular}[c]{@{}c@{}}Diagram 2 \end{tabular}} & \begin{tabular}[c]{@{}l@{}}Diagram 3\end{tabular} & \begin{tabular}[c]{@{}l@{}}All Diagrams\end{tabular}  \\ 

\cmidrule{1-5}
BLEU-1
& 0.27
& 0.25
& 0.41*
& 0.31*
\\
BLEU-2
& 0.24
& 0.27
& 0.44*
& 0.33*
\\
BLEU-3
& 0.15
& 0.23
& 0.39
& 0.26*
\\
BLEU-4
& 0.02
& 0.21
& 0.13
& 0.13
\\ 
\cmidrule{1-5}
SacreBleu 
& 0.24
& 0.30
& 0.40*
& 0.30*
\\ 
\cmidrule{1-5}
METEOR
& 0.11
& -0.04
&  0.16
& 0.09
\\ 
\cmidrule{1-5}
Rouge-1   
& 0.27
& 0.24
& 0.41*
& 0.29*
\\
Rouge-2
& 0.11
& 0.29
& 0.48*
& 0.29*
\\
Rouge-L   
& 0.29
& 0.28
& 0.34
& 0.29*
\\ 
\cmidrule{1-5}
BERTScore 
& {\bf 0.37}
& 0.21
& 0.52*
& 0.37*
\\ 
\cmidrule{1-5}
BLEURT    
& 0.25
& {\bf 0.38}
& {\bf 0.58*}
& {\bf 0.39*}

\end{tabular}}
 \footnotesize
  \begin{tablenotes}
    \item[Notes] { \emph Significance of correlation: * denotes p-values $<$ 0.05}
  \end{tablenotes}
  
\caption{Highest absolute Spearman correlation between automatic evaluation metrics and human ratings for Informativeness, where the bold font represents the highest correlation coefficient obtained by an automatic evaluation metric}
\label{table:Informativenesscorr}
}
\end{table}

\begin{table}[!htb]
\centering
\scalebox{0.65}{
\begin{tabular}{lllll}
Metric    & \multicolumn{1}{c}{\begin{tabular}[c]{@{}c@{}}Diagram 1 \end{tabular}} & \multicolumn{1}{c}{\begin{tabular}[c]{@{}c@{}}Diagram 2 \end{tabular}} & \begin{tabular}[c]{@{}l@{}}Diagram 3\end{tabular} & \begin{tabular}[c]{@{}l@{}}All Diagrams\end{tabular}  \\ 
\cmidrule{1-5}
BLEU-1
& 0.25
& 0.09
& 0.34
& 0.24*
\\
BLEU-2
& 0.24
& 0.15
& 0.41*
& 0.22
\\
BLEU-3
& 0.01
& 0.10
& 0.31
& 0.14
\\
BLEU-4
& -0.01
& 0.09
& 0.18 
& 0.10 
\\ 
\cmidrule{1-5}
SacreBleu
& 0.16
& 0.15
& 0.38 
& 0.23 
\\ 
\cmidrule{1-5}
METEOR
& 0.17
& 0.13
& 0.30 
& 0.21
\\ 
\cmidrule{1-5}
Rouge-1  
& 0.20
& 0.11
& 0.29 
& 0.20
\\
Rouge-2
& 0
& {\bf 0.24}
& 0.46*
& 0.22
\\
Rouge-L
& 0.21
& 0.09
& 0.33 
& 0.21
\\ 
\cmidrule{1-5}
BERTScore
& {\bf 0.33}
& 0.23
& 0.43*
& 0.33*
\\ 
\cmidrule{1-5}
BLEURT
& 0.26
& 0.22
& {\bf 0.53*}
& {\bf 0.34*}

\end{tabular}}
 \footnotesize{
  \begin{tablenotes}
    \item[Notes] { \emph Significance of correlation: * denotes p-values $<$ 0.05 \\ }
  \end{tablenotes}}
\caption{Spearman correlation between automatic evaluation metrics and human ratings for Clarity, where the bold font represents the highest correlation coefficient obtained by an automatic evaluation metric}
\label{table:Claritycorr}
\end{table}

\begin{table*}
\centering
\scalebox{0.58}{
\begin{tabular}{|l|l|l|l|l|l|l|c|c|c|c|c|c|c|}
\hline
\rowcolor[HTML]{C6EFCE} 
{\color[HTML]{000000} \textbf{Good Explanations}}                                                          & {\color[HTML]{000000} B1}   & {\color[HTML]{000000} B2}   & {\color[HTML]{000000} B3}   & {\color[HTML]{000000} B4}   & {\color[HTML]{000000} SB}   & {\color[HTML]{000000} M}    & \multicolumn{1}{l|}{\cellcolor[HTML]{C6EFCE}{\color[HTML]{000000} R1}} & \multicolumn{1}{l|}{\cellcolor[HTML]{C6EFCE}{\color[HTML]{000000} R2}} & \multicolumn{1}{l|}{\cellcolor[HTML]{C6EFCE}{\color[HTML]{000000} RL}} & \multicolumn{1}{l|}{\cellcolor[HTML]{C6EFCE}{\color[HTML]{000000} BS}} & \multicolumn{1}{l|}{\cellcolor[HTML]{C6EFCE}{\color[HTML]{000000} BRT}} & {\color[HTML]{000000} \begin{tabular}[c]{@{}c@{}}Inf.\end{tabular}}                         & {\color[HTML]{000000} \begin{tabular}[c]{@{}c@{}}Clar.\end{tabular}}                         \\ \hline
\rowcolor[HTML]{FFFFFF} 
{\color[HTML]{000000}(1) The alarm is triggered by a burglary or an earthquake.}                                                                                                                                                                                                                         & {\color[HTML]{000000} 0.19} & {\color[HTML]{000000} 0.12} & {\color[HTML]{000000} 0.00} & {\color[HTML]{000000} 0.00} & {\color[HTML]{000000} 0.05} & {\color[HTML]{000000} 0.23} & {\color[HTML]{000000} 0.25}                                            & {\color[HTML]{000000} 0.09}                                            & {\color[HTML]{000000} 0.12}                                            & {\color[HTML]{000000} 0.51}                                            & {\color[HTML]{000000} 0.52}                                             & {\color[HTML]{000000} 7}                                                                           & {\color[HTML]{000000} 7}                                                                            \\ \hline
\rowcolor[HTML]{FFFFFF} 
{\color[HTML]{000000} \begin{tabular}[c]{@{}l@{}}(2) Cloudy weather may produce rain and activation of the sprinkler.\\Both rain and sprinkler activity makes the grass wet.\end{tabular}}                                                                                                                                                                                                                                                                                                                         & {\color[HTML]{000000} 0.28} & {\color[HTML]{000000} 0.11} & {\color[HTML]{000000} 0.00} & {\color[HTML]{000000} 0.00} & {\color[HTML]{000000} 0.05} & {\color[HTML]{000000} 0.15} & {\color[HTML]{000000} 0.36}                                            & {\color[HTML]{000000} 0.10}                                            & {\color[HTML]{000000} 0.28}                                            & {\color[HTML]{000000} 0.49}                                            & {\color[HTML]{000000} 0.65}                                             & {\color[HTML]{000000} 7}                                                                           & {\color[HTML]{000000} 7}                                                                            \\ \hline

\rowcolor[HTML]{FFFFFF} 
\color[HTML]{000000} \begin{tabular}[c]{@{}l@{}}(3) Cost is dictated by the harvest (e.g. size) and \\ available subsidies (e.g. government tax break/subsidy). \\Whether or not the product is bought depends on the cost.\end{tabular}                                                                                                                                                                                                                                                                              & {\color[HTML]{000000} 0.18} & {\color[HTML]{000000} 0.09} & {\color[HTML]{000000} 0.00} & {\color[HTML]{000000} 0.00} & {\color[HTML]{000000} 0.02} & {\color[HTML]{000000} 0.07} & {\color[HTML]{000000} 0.25}                                            & {\color[HTML]{000000} 0.09}                                            & {\color[HTML]{000000} 0.12}                                            & {\color[HTML]{000000} 0.20}                                            & {\color[HTML]{000000} 0.51}                                             & {\color[HTML]{000000} 7}                                                                           & {\color[HTML]{000000} 7}                                                                            \\ \hline

\rowcolor[HTML]{FFC7CE} 
{\color[HTML]{000000} \textbf{Bad Explanations}}                                                                                                                                                                                                                                                                                                                                                                                                                                                                            & {\color[HTML]{000000} B1}   & {\color[HTML]{000000} B2}   & {\color[HTML]{000000} B3}   & {\color[HTML]{000000} B4}   & {\color[HTML]{000000} SB}   & {\color[HTML]{000000} M}    & {\color[HTML]{000000} R1}                                              & {\color[HTML]{000000} R2}                                              & {\color[HTML]{000000} RL}                                              & {\color[HTML]{000000} BS}                                              & {\color[HTML]{000000} BRT}                                              & \cellcolor[HTML]{FFC7CE}{\color[HTML]{000000} \begin{tabular}[c]{@{}c@{}}Inf.\end{tabular}} & \cellcolor[HTML]{FFC7CE}{\color[HTML]{000000} \begin{tabular}[c]{@{}c@{}}Clar.\end{tabular}} \\ \hline
\rowcolor[HTML]{FFFFFF} 
{\color[HTML]{000000} (4) Sensors = Alarm = prevention or ALERT.}
& {\color[HTML]{000000} 0.06} & {\color[HTML]{000000} 0.00} & {\color[HTML]{000000} 0.00} & {\color[HTML]{000000} 0.00} & {\color[HTML]{000000} 0.01} & {\color[HTML]{000000} 0.04} & {\color[HTML]{000000} 0.00}                                            & {\color[HTML]{000000} 0.00}                                            & {\color[HTML]{000000} 0.04}                                            & {\color[HTML]{000000} 0.00}                                            & {\color[HTML]{000000} 0.00}                                             & {\color[HTML]{000000} 1}                                                                           & {\color[HTML]{000000} 1}                                                                            \\ \hline

\rowcolor[HTML]{FFFFFF} 
{\color[HTML]{000000} (5) A diagram detailing a system whose goal is to make grass wet.}                                                                                                                                                                                                                                                                                                                                                                                                                               & {\color[HTML]{000000} 0.08} & {\color[HTML]{000000} 0.00} & {\color[HTML]{000000} 0.00} & {\color[HTML]{000000} 0.00} & {\color[HTML]{000000} 0.02} & {\color[HTML]{000000} 0.00} & {\color[HTML]{000000} 0.11}                                            & {\color[HTML]{000000} 0.00}                                            & {\color[HTML]{000000} 0.12}                                            & {\color[HTML]{000000} 0.13}                                            & {\color[HTML]{000000} 0.00}                                             & {\color[HTML]{000000} 1.5}                                                                         & {\color[HTML]{000000} 2}                                                                            \\ \hline

\rowcolor[HTML]{FFFFFF} 
{\color[HTML]{000000} \begin{tabular}[c]{@{}l@{}}{(6) The harvest and subsidy contribute to the cost, cost then buys??}\end{tabular}}                                                                                                                                                                                                                                                                                                                                                                                 & {\color[HTML]{000000} 0.28} & {\color[HTML]{000000} 0.00} & {\color[HTML]{000000} 0.00} & {\color[HTML]{000000} 0.00} & {\color[HTML]{000000} 0.03} & {\color[HTML]{000000} 0.15} & {\color[HTML]{000000} 0.36}                                            & {\color[HTML]{000000} 0.00}                                            & {\color[HTML]{000000} 0.24}                                            & {\color[HTML]{000000} 0.20}                                            & {\color[HTML]{000000} 0.30}                                             & {\color[HTML]{000000} 2}                                                                           & {\color[HTML]{000000} 1.5}                                                                          \\ \hline

\rowcolor[HTML]{FFEB9C} 
{\color[HTML]{000000} \bf Mixed Explanations}
& {\color[HTML]{000000} B1}   
& {\color[HTML]{000000} B2}   
& {\color[HTML]{000000} B3}   
& {\color[HTML]{000000} B4}   
& {\color[HTML]{000000} SB}   
& {\color[HTML]{000000} M}    
& {\color[HTML]{000000} R1}
& {\color[HTML]{000000} R2}
& {\color[HTML]{000000} RL}
& {\color[HTML]{000000} BS}
& {\color[HTML]{000000} BRT}
& \cellcolor[HTML]{FFEB9C}{\color[HTML]{000000} \begin{tabular}[c]{@{}c@{}}Inf.\end{tabular}} 
& \cellcolor[HTML]{FFEB9C}{\color[HTML]{000000} \begin{tabular}[c]{@{}c@{}}Clar.\end{tabular}} 
\\ \hline

\rowcolor[HTML]{FFFFFF} 
{\color[HTML]{000000} (7) The grass is getting wet.}
& {\color[HTML]{000000} 0.08} 
& {\color[HTML]{000000} 0.00} 
& {\color[HTML]{000000} 0.00} 
& {\color[HTML]{000000} 0.00} 
& {\color[HTML]{000000} 0.00} 
& {\color[HTML]{000000} 0.20} 
& {\color[HTML]{000000} 0.13}
& {\color[HTML]{000000} 0.00}                                            
& {\color[HTML]{000000} 0.15}
& {\color[HTML]{000000} 0.24}                                            
& {\color[HTML]{000000} 0.16}                                             
& {\color[HTML]{000000} 1.5}
& {\color[HTML]{000000} 7}
\\ \hline

\rowcolor[HTML]{FFFFFF} 
{\color[HTML]{000000} (8) Subsidy and harvest independently affect cost. Cost affects buys.}
& {\color[HTML]{000000} 0.06} 
& {\color[HTML]{000000} 0.00} 
& {\color[HTML]{000000} 0.00} 
& {\color[HTML]{000000} 0.00} 
& {\color[HTML]{000000} 0.01} 
& {\color[HTML]{000000} 0.14} 
& {\color[HTML]{000000} 0.16}                                            
& {\color[HTML]{000000} 0.00}                                            
& {\color[HTML]{000000} 0.10}                                            
& {\color[HTML]{000000} 0.25}                                            
& {\color[HTML]{000000} 0.56}                                             
& {\color[HTML]{000000} 6}
& {\color[HTML]{000000} 2.5}
\\ \hline

\rowcolor[HTML]{FFFFFF} 
{\color[HTML]{000000} \begin{tabular}[c]{@{}l@{}} (9) Cloud cover influences whether it rains and when the sprinkler is activated. \\ When either the sprinkler is turned on or when it rains, the grass gets wet. \end{tabular}}      
& {\color[HTML]{000000} 0.48} 
& {\color[HTML]{000000} 0.33} 
& {\color[HTML]{000000} 0.21} 
& {\color[HTML]{000000} 0.14} 
& {\color[HTML]{000000} 0.22} 
& {\color[HTML]{000000} 0.24} 
& {\color[HTML]{000000} 0.50}                                            
& {\color[HTML]{000000} 0.24}                                            
& {\color[HTML]{000000} 0.38}                                            
& {\color[HTML]{000000} 0.49}                                            
& {\color[HTML]{000000} 0.65}                                             
& {\color[HTML]{000000} 7}                                                   &  {\color[HTML]{000000} 3}
\\ \hline

\end{tabular}}
\caption{Examples of Good, Bad and Mixed Explanations according to human evaluation scores for Informativeness and Clarity (medians of all ratings for that explanation), presented with their automatic measures}
\label{table:examples}
\vspace{-0.3cm}
\end{table*}

\subsection{Discussion}
BLEU-based metrics can be easily and quickly computed; however, they do not correlate as well with human ratings as other methods presented here. This might be due to certain limitations, such as the fact that they rely on word overlap and are not invariant to paraphrases. Furthermore, they do not use recall, rather a Brevity Penalty, which penalizes generated text for being \say{too short} \citep{Papineni2001}. This way may not be appropriate for explanations, as good explanations may need to be lengthy by their very nature. 

METEOR takes into consideration F1-measure by computing scores for unigram precision and recall. The fragmentation penalty is calculated using the total number of matched words (m, averaged over hypothesis and reference) and the number of chunks. In this way, it could identify synonyms, but perhaps not as well as the embedding-based metrics, as evidenced by the correlation figures in our results. With regards to ROUGE-based scores, due to the upper bound issues presented by \newcite{Schluter2017}, it is impossible to obtain perfect ROUGE-n scores. Furthermore, ROUGE-L cannot differentiate if the reference and the candidate have the same longest common subsequence (LCS), but different word ordering. Again, word ordering may be important for the explanation in terms of explainee scaffolding \cite{palincsar}. 
 
 It has been brought into question whether a single automatic measure is able to capture multiple aspects of subjective human evaluation \cite{belz-etal-2020-disentangling}. 
  Thus, in order to understand to what degree the various metrics capture both Clarity and Informativeness, we investigated individual explanations and their ratings. Table \ref{table:examples} gives some extracts from the dataset along with the automatic metrics and the human evaluation scores of Informativeness and Clarity. Based on these human scores, the extracts are divided into:
 good explanations (high scores for both), bad explanations (low scores for both) and mixed explanations (mixed scores). We can see here that all metrics are reasonably good at capturing and evaluating the `bad' explanations with low scores across the board. However, only the BLEURT metric is good at capturing both `good and bad' explanation ratings, as observed in the difference in scores between these two categories. ROUGE-L and BERTScore do capture this difference in some cases, but they are not as consistent as BLEURT. The reason that BLEURT outperforms the other metrics may be because it uses a combination of word-overlap metrics as well as embeddings and thus may be capturing the best of these approaches. 
\newpage
 Although Clarity and Informativeness highly correlate overall, there are occasions where explanations are rated by humans as higher on Clarity than Informativeness and visa-versa. However, there are rarely any cases where Clarity is high, and Informativeness is very low. Explanation 8 in Table \ref{table:examples} is the only example of this in our corpus. It is thus difficult to make any generalisations about this subset of the data. However, it does seem to be the case that BLEURT is more sensitive to Informativeness than Clarity (e.g. explanation 7 vs 8-9 in the table), but a larger study would be needed to show this empirically.

\section{Conclusions and Future work}
 
  Human evaluation is an expensive and time-consuming process. On the other hand, automatic evaluation is a cheaper and more efficient method for evaluating NLG systems. However, finding accurate measures is challenging, particularly for explanations. We have discussed word embedding techniques \cite{Mikolov2013, kim-2014-convolutional, Reimers2020}, which enable the use of pre-trained models and so reduces the need to collect large amounts of data in our domain of explanations, which is a challenging task. The embedding-based metrics mentioned here perform better than the word-overlap based ones. We speculate that this is in part due to the fact that the former capture semantics more effectively and are thus more invariant to paraphrases. These metrics have also been shown to be useful across multiple tasks \cite{sellam-etal-2020-bleurt}
  but with some variation across datasets \cite{novikova-etal-2017-need}. Therefore, future work would involve examining the effectiveness of automatic metrics across a wider variety of explanation tasks and datasets, as outlined in the Related Work section.  
 
Embeddings are quite opaque in themselves. Whilst some attempts have been made to visualise them \cite{li-etal-2016-visualizing}, it remains that embedding-based metrics do not provide much insight into what makes a good/bad explanation. It would thus be necessary to look more deeply into the linguistic phenomena that may indicate the quality of explanations. In ExBAN, initial findings show that the number of nouns and coordinating conjunctions correlate with human judgements, however further in-depth analysis is needed. Additional metrics to add to the set explored here could include grammar-based metrics, such as readability and grammaticality, as in the study described in  \cite{novikova-etal-2017-need}.
 
 Furthermore, an investigation is needed into the pragmatic and cognitive processes underlying explanations, such as argumentation, reasoning, causality, and common sense \cite{Baaj2019}. Investigating whether these can be captured automatically will be highly challenging.  We will explore further the idea of adapting explanations to the explainee's knowledge and expertise level, as well as the explainer's goals and intentions. One such goal of the explainer could be to maximise the trustworthiness of the explanation \cite{Ribeiro2016}. How this aspect is consistently subjectively and objectively measured will be an interesting area of investigation.

 Finally, the ExBAN corpus and this study will inform the development of NLG algorithms for NL explanations from graphical representations. We will explore NLG techniques for structured data, such as graph neural networks and knowledge graphs \cite{Kedziorski}. Thus the corpus and metrics discussed here will contribute to a variety of fields linguistics, cognitive science as well as NLG and Explainable AI.
 
\section*{Acknowledgments}

This work was supported by the EPSRC Centre for Doctoral Training in Robotics and Autonomous Systems at Heriot-Watt University and the University of Edinburgh. Clinciu's PhD is funded by Schlumberger Cambridge Research Limited (EP/L016834/1, 2018-2021). This work was also supported by the EPSRC ORCA Hub (EP/R026173/1, 2017-2021) and UKRI Trustworthy Autonomous Systems Node on Trust (EP/V026682/1, 2020-2024).

\balance
\bibliography{eacl2021}

\begin{thebibliography}{82}
\expandafter\ifx\csname natexlab\endcsname\relax\def\natexlab#1{#1}\fi

\bibitem[{Arya et~al.(2019)Arya, Bellamy, Chen, Dhurandhar, Hind, Hoffman,
  Houde, Liao, Luss, Mojsilovic, Mourad, Pedemonte, Raghavendra, Richards,
  Sattigeri, Shanmugam, Singh, Varshney, Wei, and Zhang}]{aix360-sept-2019}
Vijay Arya, Rachel K.~E. Bellamy, Pin{-}Yu Chen, Amit Dhurandhar, Michael Hind,
  Samuel~C. Hoffman, Stephanie Houde, Q.~Vera Liao, Ronny Luss, Aleksandra
  Mojsilovic, Sami Mourad, Pablo Pedemonte, Ramya Raghavendra, John~T.
  Richards, Prasanna Sattigeri, Karthikeyan Shanmugam, Moninder Singh, Kush~R.
  Varshney, Dennis Wei, and Yunfeng Zhang. 2019.
\newblock \href {http://arxiv.org/abs/1909.03012} {One explanation does not fit
  all: {A} toolkit and taxonomy of {AI} explainability techniques}.
\newblock \emph{CoRR}, abs/1909.03012.

\bibitem[{Baaj et~al.(2019)Baaj, Poli, and Ouerdane}]{Baaj2019}
Isma{\"\i}l Baaj, Jean-Philippe Poli, and Wassila Ouerdane. 2019.
\newblock \href {https://doi.org/10.18653/v1/W19-8404} {Some insights towards a
  unified semantic representation of explanation for e{X}plainable artificial
  intelligence}.
\newblock In \emph{Proceedings of the 1st Workshop on Interactive Natural
  Language Technology for Explainable Artificial Intelligence (NL4XAI 2019)},
  pages 14--19. Association for Computational Linguistics.

\bibitem[{Belz and Kow(2009)}]{Belz2009kow}
Anja Belz and Eric Kow. 2009.
\newblock \href {https://doi.org/10.3115/1610195.1610198} {{System building
  cost vs. output quality in data-to-text generation}}.
\newblock In \emph{Proceedings of the 12th European Workshop on Natural
  Language Generation, ENLG 2009}.

\bibitem[{Belz and Reiter(2006)}]{Belz2006}
Anja Belz and Ehud Reiter. 2006.
\newblock \href {https://www.aclweb.org/anthology/E06-1040} {Comparing
  automatic and human evaluation of {NLG} systems}.
\newblock In \emph{Proceedings of the 11th Conference of the {E}uropean Chapter
  of the Association for Computational Linguistics}, Trento, Italy. Association
  for Computational Linguistics.

\bibitem[{Belz et~al.(2020)Belz, Mille, and
  Howcroft}]{belz-etal-2020-disentangling}
Anya Belz, Simon Mille, and David~M. Howcroft. 2020.
\newblock \href {https://www.aclweb.org/anthology/2020.inlg-1.24}
  {Disentangling the properties of human evaluation methods: A classification
  system to support comparability, meta-evaluation and reproducibility
  testing}.
\newblock In \emph{Proceedings of the 13th International Conference on Natural
  Language Generation}, pages 183--194, Dublin, Ireland. Association for
  Computational Linguistics.

\bibitem[{Bowman et~al.(2015)Bowman, Angeli, Potts, and Manning}]{Bowman2015}
Samuel~R. Bowman, Gabor Angeli, Christopher Potts, and Christopher~D. Manning.
  2015.
\newblock \href {https://doi.org/10.18653/v1/D15-1075} {A large annotated
  corpus for learning natural language inference}.
\newblock In \emph{Proceedings of the 2015 Conference on Empirical Methods in
  Natural Language Processing}, pages 632--642, Lisbon, Portugal. Association
  for Computational Linguistics.

\bibitem[{Braun et~al.(2018)Braun, Reiter, and Siddharthan}]{Braun2018}
Daniel Braun, Ehud Reiter, and Advaith Siddharthan. 2018.
\newblock \href {https://doi.org/10.1017/S1351324918000050} {{SaferDrive: An
  NLG-based behaviour change support system for drivers}}.
\newblock \emph{Natural Language Engineering}, 24(4).

\bibitem[{Cai et~al.(2019)Cai, Jongejan, and Holbrook}]{Cai2019}
Carrie~J. Cai, Jonas Jongejan, and Jess Holbrook. 2019.
\newblock \href {https://doi.org/10.1145/3301275.3302289} {{The effects of
  example-based explanations in a machine learning interface}}.
\newblock In \emph{International Conference on Intelligent User Interfaces,
  Proceedings IUI}, volume Part F147615.

\bibitem[{Camburu et~al.(2018)Camburu, Rockt\"{a}schel, Lukasiewicz, and
  Blunsom}]{Camburu2018}
Oana-Maria Camburu, Tim Rockt\"{a}schel, Thomas Lukasiewicz, and Phil Blunsom.
  2018.
\newblock \href
  {https://proceedings.neurips.cc/paper/2018/file/4c7a167bb329bd92580a99ce422d6fa6-Paper.pdf}
  {e-snli: Natural language inference with natural language explanations}.
\newblock In \emph{Advances in Neural Information Processing Systems},
  volume~31, pages 9539--9549. Curran Associates, Inc.

\bibitem[{Chiyah~Garcia et~al.(2018)Chiyah~Garcia, Robb, Liu, Laskov, Patron,
  and Hastie}]{GarciaetalINLG18}
Francisco~Javier Chiyah~Garcia, David~A. Robb, Xingkun Liu, Atanas Laskov,
  Pedro Patron, and Helen Hastie. 2018.
\newblock \href {https://doi.org/10.18653/v1/W18-6511} {Explainable autonomy: A
  study of explanation styles for building clear mental models}.
\newblock In \emph{Proceedings of the 11th International Conference on Natural
  Language Generation}, pages 99--108, Tilburg University, The Netherlands.
  Association for Computational Linguistics.

\bibitem[{Clinciu and Hastie(2019)}]{Clinciu2019}
Miruna-Adriana Clinciu and Helen Hastie. 2019.
\newblock \href {https://doi.org/10.18653/v1/w19-8403} {{A Survey of
  Explainable AI Terminology}}.
\newblock \emph{Proceedings of the 1st Workshop on Interactive Natural Language
  Technology for Explainable Artificial Intelligence (NL4XAI 2019)}, pages
  8--13.

\bibitem[{Commission(2018)}]{gdprart22}
European Commission. 2018.
\newblock {A}rticle 22 {EU} {GDPR} "{A}utomated individual decision-making,
  including profiling".
\newblock \url{https://www.privacy-regulation.eu/en/22.htm}.
\newblock Accessed on 2021-01-25.

\bibitem[{Dali et~al.(2009)Dali, Rusu, Fortuna, Mladeni{\'{c}}, and
  Grobelnik}]{Dali2009}
Lorand Dali, Delia Rusu, Bla{\v{z}} Fortuna, Dunja Mladeni{\'{c}}, and Marko
  Grobelnik. 2009.
\newblock {Question answering based on semantic graphs}.
\newblock In \emph{CEUR Workshop Proceedings}, volume 491.

\bibitem[{{De Graaf} and Malle(2017)}]{DeGraaf2017}
Maartje~M.A. {De Graaf} and Bertram~F. Malle. 2017.
\newblock {How people explain action (and autonomous intelligent systems should
  too)}.
\newblock In \emph{AAAI Fall Symposium - Technical Report}, volume FS-17-01 -
  FS-17-05.

\bibitem[{Dethlefs et~al.(2014)Dethlefs, Cuay{\'{a}}huitl, Hastie, Rieser, and
  Lemon}]{Dethlefs2014}
Nina Dethlefs, Heriberto Cuay{\'{a}}huitl, Helen Hastie, Verena Rieser, and
  Oliver Lemon. 2014.
\newblock {Cluster-based prediction of user ratings for stylistic surface
  realisation}.
\newblock In \emph{Proceedings of the 14th Conference of the European Chapter
  of the Association for Computational Linguistics 2014, EACL 2014}, pages
  702--711. Association for Computational Linguistics (ACL).

\bibitem[{Deutch et~al.(2016)Deutch, Frost, and Gilad}]{Deutch2015}
Daniel Deutch, Nave Frost, and Amir Gilad. 2016.
\newblock \href {https://doi.org/10.14778/3007263.3007303} {Nlprov: Natural
  language provenance}.
\newblock In \emph{Proceedings of the 42nd International Conference on Very
  Large Data Bases (VLDB) Endowment}, volume~9, page 1537–1540. VLDB
  Endowment.

\bibitem[{Devlin et~al.(2019)Devlin, Chang, Lee, and Toutanova}]{Devlin2019}
Jacob Devlin, Ming-Wei Chang, Kenton Lee, and Kristina Toutanova. 2019.
\newblock \href {https://doi.org/10.18653/v1/N19-1423} {{BERT}: Pre-training of
  deep bidirectional transformers for language understanding}.
\newblock In \emph{Proceedings of the 2019 Conference of the North {A}merican
  Chapter of the Association for Computational Linguistics: Human Language
  Technologies, Volume 1 (Long and Short Papers)}, pages 4171--4186,
  Minneapolis, Minnesota. Association for Computational Linguistics.

\bibitem[{Doshi-Velez and Kim(2017)}]{Doshi-Velez2017}
Finale Doshi-Velez and B.~Kim. 2017.
\newblock Towards a rigorous science of interpretable machine learning.
\newblock \emph{arXiv: Machine Learning}.

\bibitem[{Du{\v{s}}ek et~al.(2020)Du{\v{s}}ek, Novikova, and
  Rieser}]{Dusek2020}
Ond\v{r}ej Du{\v{s}}ek, Jekaterina Novikova, and Verena Rieser. 2020.
\newblock \href {https://doi.org/10.1016/j.csl.2019.06.009} {Evaluating the
  {{State}}-of-the-{{Art}} of {{End}}-to-{{End Natural Language Generation}}:
  {{The E2E NLG Challenge}}}.
\newblock \emph{Computer Speech \& Language}, 59:123--156.

\bibitem[{Fan et~al.(2019)Fan, Jernite, Perez, Grangier, Weston, and
  Auli}]{fan2019eli5}
Angela Fan, Yacine Jernite, Ethan Perez, David Grangier, Jason Weston, and
  Michael Auli. 2019.
\newblock \href {https://doi.org/10.18653/v1/P19-1346} {{ELI}5: Long form
  question answering}.
\newblock In \emph{Proceedings of the 57th Annual Meeting of the Association
  for Computational Linguistics}, pages 3558--3567, Florence, Italy.
  Association for Computational Linguistics.

\bibitem[{Gal and Ghahramani(2016)}]{Gal2016}
Yarin Gal and Zoubin Ghahramani. 2016.
\newblock \href {http://proceedings.mlr.press/v48/gal16.html} {Dropout as a
  bayesian approximation: Representing model uncertainty in deep learning}.
\newblock In \emph{Proceedings of The 33rd International Conference on Machine
  Learning}, volume~48 of \emph{Proceedings of Machine Learning Research},
  pages 1050--1059, New York, New York, USA. PMLR.

\bibitem[{Gatt and Krahmer(2018)}]{Gatt2018}
Albert Gatt and Emiel Krahmer. 2018.
\newblock \href {https://doi.org/10.1613/jair.5714} {{Survey of the state of
  the art in natural language generation: Core tasks, applications and
  evaluation}}.
\newblock \emph{Journal of Artificial Intelligence Research}, 61:1--64.

\bibitem[{Gkatzia and Mahamood(2015)}]{gkatzia-mahamood-2015-snapshot}
Dimitra Gkatzia and Saad Mahamood. 2015.
\newblock \href {https://doi.org/10.18653/v1/W15-4708} {A snapshot of {NLG}
  evaluation practices 2005 - 2014}.
\newblock In \emph{Proceedings of the 15th {E}uropean Workshop on Natural
  Language Generation ({ENLG})}, pages 57--60, Brighton, UK. Association for
  Computational Linguistics.

\bibitem[{Goodrich et~al.(2019)Goodrich, Rao, Liu, and Saleh}]{Goodrich2019}
Ben Goodrich, Vinay Rao, Peter~J. Liu, and Mohammad Saleh. 2019.
\newblock \href {https://doi.org/10.1145/3292500.3330955} {{Assessing the
  factual accuracy of generated text}}.
\newblock In \emph{Proceedings of the ACM SIGKDD International Conference on
  Knowledge Discovery and Data Mining}.

\bibitem[{Gregor and Benbasat(1999)}]{Gregor1999}
Shirley Gregor and Izak Benbasat. 1999.
\newblock \href {https://doi.org/10.2307/249487} {{Explanations from
  intelligent systems: Theoretical foundations and implications for practice}}.
\newblock \emph{MIS Quarterly: Management Information Systems}, 23(4).

\bibitem[{Hallgren(2012)}]{Hallgren2012}
Kevin~A. Hallgren. 2012.
\newblock \href {https://doi.org/10.20982/tqmp.08.1.p023} {{Computing
  Inter-Rater Reliability for Observational Data: An Overview and Tutorial}}.
\newblock \emph{Tutorials in Quantitative Methods for Psychology}, 8(1).

\bibitem[{Harbers et~al.(2009)Harbers, {Van Den Bosch}, and
  Meyer}]{Harbers2009}
Maaike Harbers, Karel {Van Den Bosch}, and John Jules~Ch Meyer. 2009.
\newblock \href {https://doi.org/10.1007/978-3-642-04380-2_17} {{A study into
  preferred explanations of virtual agent behavior}}.
\newblock In \emph{Lecture Notes in Computer Science (including subseries
  Lecture Notes in Artificial Intelligence and Lecture Notes in
  Bioinformatics)}, volume 5773 LNAI.

\bibitem[{Hardcastle and Scott(2008)}]{Hardcastle2008}
David Hardcastle and Donia Scott. 2008.
\newblock \href
  {http://www.lrec-conf.org/proceedings/lrec2008/pdf/797_paper.pdf} {Can we
  evaluate the quality of generated text?}
\newblock In \emph{Proceedings of the Sixth International Conference on
  Language Resources and Evaluation ({LREC}'08)}, Marrakech, Morocco. European
  Language Resources Association (ELRA).

\bibitem[{Hastie and Belz(2014)}]{Hastie2014}
Helen Hastie and Anja Belz. 2014.
\newblock {A comparative evaluation methodology for NLG in interactive
  systems}.
\newblock In \emph{Proceedings of the 9th International Conference on Language
  Resources and Evaluation, LREC 2014}.

\bibitem[{Howcroft et~al.(2020)Howcroft, Belz, Clinciu, Gkatzia, Hasan,
  Mahamood, Mille, van Miltenburg, Santhanam, and
  Rieser}]{howcroft-etal-2020-twenty}
David~M. Howcroft, Anya Belz, Miruna-Adriana Clinciu, Dimitra Gkatzia, Sadid~A.
  Hasan, Saad Mahamood, Simon Mille, Emiel van Miltenburg, Sashank Santhanam,
  and Verena Rieser. 2020.
\newblock \href {https://www.aclweb.org/anthology/2020.inlg-1.23} {Twenty years
  of confusion in human evaluation: {NLG} needs evaluation sheets and
  standardised definitions}.
\newblock In \emph{Proceedings of the 13th International Conference on Natural
  Language Generation}, pages 169--182, Dublin, Ireland. Association for
  Computational Linguistics.

\bibitem[{Jansen et~al.(2019)Jansen, Wainwright, Marmorstein, and
  Morrison}]{Jansen2019}
Peter~A. Jansen, Elizabeth Wainwright, Steven Marmorstein, and Clayton~T.
  Morrison. 2019.
\newblock \href {http://arxiv.org/abs/1802.03052} {{WorldTree: A corpus of
  explanation graphs for elementary science questions supporting multi-hop
  inference}}.
\newblock In \emph{Proceedings of the 11th International Conference on Language
  Resources and Evaluation (LREC)}, pages 2732--2740. European Language
  Resources Association (ELRA).

\bibitem[{Kim(2014)}]{kim-2014-convolutional}
Yoon Kim. 2014.
\newblock \href {https://doi.org/10.3115/v1/D14-1181} {Convolutional neural
  networks for sentence classification}.
\newblock In \emph{Proceedings of the 2014 Conference on Empirical Methods in
  Natural Language Processing ({EMNLP})}, pages 1746--1751, Doha, Qatar.
  Association for Computational Linguistics.

\bibitem[{Koncel-Kedziorski et~al.(2019)Koncel-Kedziorski, Bekal, Luan, Lapata,
  and Hajishirzi}]{Kedziorski}
Rik Koncel-Kedziorski, Dhanush Bekal, Yi~Luan, Mirella Lapata, and Hannaneh
  Hajishirzi. 2019.
\newblock \href {https://doi.org/10.18653/v1/N19-1238} {{T}ext {G}eneration
  from {K}nowledge {G}raphs with {G}raph {T}ransformers}.
\newblock In \emph{Proceedings of the 2019 Conference of the North {A}merican
  Chapter of the Association for Computational Linguistics: Human Language
  Technologies, Volume 1 (Long and Short Papers)}, pages 2284--2293,
  Minneapolis, Minnesota. Association for Computational Linguistics.

\bibitem[{Kov{\'{a}}ř et~al.(2016)Kov{\'{a}}ř, Jakub{\'{i}}{\v{c}}ek, and
  Hor{\'{a}}k}]{Kovar2016}
Vojt{\v{e}}ch Kov{\'{a}}ř, Milo{\v{s}} Jakub{\'{i}}{\v{c}}ek, and Ale{\v{s}}
  Hor{\'{a}}k. 2016.
\newblock \href {https://doi.org/10.5220/0005824805400545} {{On evaluation of
  natural language processing tasks: Is gold standard evaluation methodology a
  good solution?}}
\newblock In \emph{ICAART 2016 - Proceedings of the 8th International
  Conference on Agents and Artificial Intelligence}, volume~2, pages 540--545.
  SciTePress.

\bibitem[{Krening et~al.(2017)Krening, Harrison, Feigh, Isbell, Riedl, and
  Thomaz}]{krening2017}
Samantha Krening, Brent Harrison, Karen~M. Feigh, Charles~Lee Isbell, Mark
  Riedl, and Andrea Thomaz. 2017.
\newblock \href {https://doi.org/10.1109/TCDS.2016.2628365} {{Learning From
  Explanations Using Sentiment and Advice in RL}}.
\newblock \emph{IEEE Transactions on Cognitive and Developmental Systems},
  9(1).

\bibitem[{Krippendorff(1980)}]{Krippendorff1980}
Klaus Krippendorff. 1980.
\newblock \href {https://doi.org/10.2307/2288384} {\emph{{Metodolog{\'{i}}a de
  an{\'{a}}lisis de contenido. Teor{\'{i}}a y pr{\'{a}}ctica.}}}
\newblock SAGE, 2004.

\bibitem[{Kumar and Talukdar(2020)}]{Kumar2020}
Sawan Kumar and Partha Talukdar. 2020.
\newblock \href {https://doi.org/10.18653/v1/2020.acl-main.771} {{NILE} :
  Natural language inference with faithful natural language explanations}.
\newblock In \emph{Proceedings of the 58th Annual Meeting of the Association
  for Computational Linguistics}, pages 8730--8742, Online. Association for
  Computational Linguistics.

\bibitem[{Lamm et~al.(2020)Lamm, Palomaki, Alberti, Andor, Choi, Soares, and
  Collins}]{lamm2020qed}
Matthew Lamm, Jennimaria Palomaki, Chris Alberti, Daniel Andor, Eunsol Choi,
  Livio~Baldini Soares, and Michael Collins. 2020.
\newblock \href {http://arxiv.org/abs/2009.06354} {Qed: A framework and dataset
  for explanations in question answering}.

\bibitem[{Lampouras and Androutsopoulos(2013)}]{Lampouras2013}
Gerasimos Lampouras and Ion Androutsopoulos. 2013.
\newblock \href {https://www.aclweb.org/anthology/W13-2106} {Using integer
  linear programming for content selection, lexicalization, and aggregation to
  produce compact texts from {OWL} ontologies}.
\newblock In \emph{Proceedings of the 14th {E}uropean Workshop on Natural
  Language Generation}, pages 51--60, Sofia, Bulgaria. Association for
  Computational Linguistics.

\bibitem[{Lavie and Agarwal(2007)}]{lavie-agarwal-2007-meteor}
Alon Lavie and Abhaya Agarwal. 2007.
\newblock \href {https://www.aclweb.org/anthology/W07-0734} {{METEOR}: An
  automatic metric for {MT} evaluation with high levels of correlation with
  human judgments}.
\newblock In \emph{Proceedings of the Second Workshop on Statistical Machine
  Translation}, pages 228--231, Prague, Czech Republic. Association for
  Computational Linguistics.

\bibitem[{Leake(2014)}]{Leake2014}
David~B. Leake. 2014.
\newblock \href {https://doi.org/10.4324/9781315807072} {\emph{{Evaluating
  Explanations}}}.
\newblock Psychology Press.

\bibitem[{van~der Lee et~al.(2017)van~der Lee, Krahmer, and
  Wubben}]{VanDerLee2017}
Chris van~der Lee, Emiel Krahmer, and Sander Wubben. 2017.
\newblock \href {https://doi.org/10.18653/v1/W17-3513} {{PASS}: A {D}utch
  data-to-text system for soccer, targeted towards specific audiences}.
\newblock In \emph{Proceedings of the 10th International Conference on Natural
  Language Generation}, pages 95--104, Santiago de Compostela, Spain.
  Association for Computational Linguistics.

\bibitem[{Li et~al.(2016)Li, Chen, Hovy, and
  Jurafsky}]{li-etal-2016-visualizing}
Jiwei Li, Xinlei Chen, Eduard Hovy, and Dan Jurafsky. 2016.
\newblock \href {https://doi.org/10.18653/v1/N16-1082} {Visualizing and
  understanding neural models in {NLP}}.
\newblock In \emph{Proceedings of the 2016 Conference of the North {A}merican
  Chapter of the Association for Computational Linguistics: Human Language
  Technologies}, pages 681--691, San Diego, California. Association for
  Computational Linguistics.

\bibitem[{Lin(1971)}]{Lin1971}
Chin-Yew Lin. 1971.
\newblock \href {https://doi.org/10.1253/jcj.34.1213} {{ROUGE: A Package for
  Automatic Evaluation of Summaries Chin-Yew}}.
\newblock \emph{Information Sciences Institute}, 34(12).

\bibitem[{Lombrozo(2007)}]{Lombrozo2007}
Tania Lombrozo. 2007.
\newblock \href {https://doi.org/10.1016/j.cogpsych.2006.09.006} {{Simplicity
  and probability in causal explanation}}.
\newblock \emph{Cognitive Psychology}, 55(3):232--257.

\bibitem[{Madumal et~al.(2019)Madumal, Sonenberg, Miller, and
  Vetere}]{Madumal2019}
Prashan Madumal, Liz Sonenberg, Tim Miller, and Frank Vetere. 2019.
\newblock {A grounded interaction protocol for explainable artificial
  intelligence}.
\newblock In \emph{Proceedings of the International Joint Conference on
  Autonomous Agents and Multiagent Systems, AAMAS}, volume~2.

\bibitem[{Mahapatra et~al.(2016)Mahapatra, Naskar, and
  Bandyopadhyay}]{Mahapatra2016}
Joy Mahapatra, Sudip~Kumar Naskar, and Sivaji Bandyopadhyay. 2016.
\newblock \href {https://doi.org/10.18653/v1/w16-6624} {{Statistical natural
  language generation from tabular non-textual data}}.
\newblock In \emph{INLG 2016 - 9th International Natural Language Generation
  Conference, Proceedings of the Conference}.

\bibitem[{Mascaro et~al.(2014)Mascaro, Nicholson, and Korb}]{Mascaro2014}
Steven Mascaro, Ann Nicholson, and Kevin Korb. 2014.
\newblock \href {https://doi.org/10.1016/j.ijar.2013.03.012} {{Anomaly
  detection in vessel tracks using Bayesian networks}}.
\newblock In \emph{International Journal of Approximate Reasoning}, volume~55.

\bibitem[{Maxwell et~al.(2017)Maxwell, Azzopardi, and Moshfeghi}]{Maxwell2017}
David Maxwell, Leif Azzopardi, and Yashar Moshfeghi. 2017.
\newblock \href {https://doi.org/10.1145/3077136.3080824} {{A study of snippet
  length and informativeness behaviour, performance and user experience}}.
\newblock In \emph{SIGIR 2017 - Proceedings of the 40th International ACM SIGIR
  Conference on Research and Development in Information Retrieval}.

\bibitem[{McCormick and Ryan(2019)}]{berttutorial}
Chris McCormick and Nick Ryan. 2019.
\newblock \href {http://www.mccormickml.com} {Bert word embeddings tutorial}.
\newblock Accessed on 2021-01-25.

\bibitem[{Mellish and Dale(1998)}]{Mellish1998}
C.~Mellish and R.~Dale. 1998.
\newblock \href {https://doi.org/10.1006/csla.1998.0106} {{Evaluation in the
  context of natural language generation}}.
\newblock \emph{Computer Speech and Language}, 12(4).

\bibitem[{Metelli and Heard(2019)}]{Metelli2019}
Silvia Metelli and Nicholas Heard. 2019.
\newblock \href {https://doi.org/10.1214/19-AOAS1286} {{On bayesian new edge
  prediction and anomaly detection in computer networks}}.
\newblock \emph{Annals of Applied Statistics}, 13(4).

\bibitem[{Mikolov et~al.(2013)Mikolov, Chen, Corrado, and Dean}]{Mikolov2013}
Tomas Mikolov, Kai Chen, Greg Corrado, and Jeffrey Dean. 2013.
\newblock {Efficient estimation of word representations in vector space}.
\newblock In \emph{1st International Conference on Learning Representations,
  ICLR 2013 - Workshop Track Proceedings}.

\bibitem[{Miller(2018)}]{DBLP:journals/corr/Miller17a}
Tim Miller. 2018.
\newblock \href {http://arxiv.org/abs/1706.07269} {{E}xplanation in
  {A}rtificial {I}ntelligence: {I}nsights from the {S}ocial {S}ciences}.
\newblock \emph{arXiv preprint arXiv:1706.07269}.

\bibitem[{Miller et~al.(2017)Miller, Hower, and Sonenberg}]{Miller2017}
Tim Miller, Piers Hower, and Liz Sonenberg. 2017.
\newblock \href {https://doi.org/10.1016/j.foodchem.2017.11.091} {{Explainable
  AI: beware of inmates running the asylum}}.
\newblock In \emph{Proceedings of the IJCAI 2017 workshop on explainable
  artificial intelligence (XAI)}, October, page 363.

\bibitem[{Novikova et~al.(2017)Novikova, Du{\v{s}}ek, Cercas~Curry, and
  Rieser}]{novikova-etal-2017-need}
Jekaterina Novikova, Ond{\v{r}}ej Du{\v{s}}ek, Amanda Cercas~Curry, and Verena
  Rieser. 2017.
\newblock \href {https://doi.org/10.18653/v1/D17-1238} {Why we need new
  evaluation metrics for {NLG}}.
\newblock In \emph{Proceedings of the 2017 Conference on Empirical Methods in
  Natural Language Processing}, pages 2241--2252, Copenhagen, Denmark.
  Association for Computational Linguistics.

\bibitem[{Novikova et~al.(2018)Novikova, Du{\v{s}}ek, and Rieser}]{Novikova}
Jekaterina Novikova, Ond{\v{r}}ej Du{\v{s}}ek, and Verena" Rieser. 2018.
\newblock \href {https://doi.org/10.18653/v1/N18-2012} {{R}ank{ME}: Reliable
  human ratings for natural language generation}.
\newblock In \emph{Proceedings of the 2018 Conference of the North {A}merican
  Chapter of the Association for Computational Linguistics: Human Language
  Technologies, Volume 2 (Short Papers)}, pages 72--78, New Orleans, Louisiana.
  Association for Computational Linguistics.

\bibitem[{Palincsar(1986)}]{palincsar}
Annemarie~Sullivan Palincsar. 1986.
\newblock \href {https://doi.org/10.1080/00461520.1986.9653025} {The role of
  dialogue in providing scaffolded instruction}.
\newblock \emph{Educational Psychologist}, 21(1-2):73--98.

\bibitem[{Papineni et~al.(2001)Papineni, Roukos, Ward, Zhu, and
  Heights}]{Papineni2001}
Kishore Papineni, Salim Roukos, Todd Ward, Wei-jing Zhu, and Yorktown Heights.
  2001.
\newblock \href {https://doi.org/10.3115/1073083.1073135} {{IBM Research Report
  Bleu : a Method for Automatic Evaluation of Machine Translation}}.
\newblock \emph{Science}, 22176:1--10.

\bibitem[{Park et~al.(2018)Park, Hendricks, Akata, Rohrbach, Schiele, Darrell,
  and Rohrbach}]{Park2018}
Dong~Huk Park, Lisa~Anne Hendricks, Zeynep Akata, Anna Rohrbach, Bernt Schiele,
  Trevor Darrell, and Marcus Rohrbach. 2018.
\newblock \href {https://doi.org/10.1109/CVPR.2018.00915} {{Multimodal
  Explanations: Justifying Decisions and Pointing to the Evidence}}.
\newblock In \emph{Proceedings of the IEEE Computer Society Conference on
  Computer Vision and Pattern Recognition}, pages 8779--8788. IEEE Computer
  Society.

\bibitem[{Plumb et~al.(2018)Plumb, Molitor, and Talwalkar}]{Plumb2018}
Gregory Plumb, Denali Molitor, and Ameet Talwalkar. 2018.
\newblock {Model agnostic supervised local explanations}.
\newblock In \emph{Advances in Neural Information Processing Systems}, volume
  2018-December.

\bibitem[{Post(2018)}]{post-2018-call}
Matt Post. 2018.
\newblock \href {https://doi.org/10.18653/v1/W18-6319} {A call for clarity in
  reporting {BLEU} scores}.
\newblock In \emph{Proceedings of the Third Conference on Machine Translation:
  Research Papers}, pages 186--191, Belgium, Brussels. Association for
  Computational Linguistics.

\bibitem[{Rajani et~al.(2019)Rajani, McCann, Xiong, and
  Socher}]{rajani2019explain}
Nazneen~Fatema Rajani, Bryan McCann, Caiming Xiong, and Richard Socher. 2019.
\newblock \href {https://arxiv.org/abs/1906.02361} {Explain yourself!
  leveraging language models for commonsense reasoning}.
\newblock In \emph{Proceedings of the 2019 Conference of the Association for
  Computational Linguistics (ACL2019)}.

\bibitem[{Reimers and Gurevych(2020)}]{Reimers2020}
Nils Reimers and Iryna Gurevych. 2020.
\newblock \href {https://doi.org/10.18653/v1/d19-1410} {{Sentence-BERT:
  Sentence embeddings using siamese BERT-networks}}.
\newblock In \emph{Proceedings of the EMNLP-IJCNLP 2019 - 2019 Conference on
  Empirical Methods in Natural Language Processing and 9th International Joint
  Conference on Natural Language Processing, Proceedings of the Conference}.

\bibitem[{Ribeiro et~al.(2016)Ribeiro, Singh, and Guestrin}]{Ribeiro2016}
Marco~Tulio Ribeiro, Sameer Singh, and Carlos Guestrin. 2016.
\newblock \href {https://doi.org/10.1145/2939672.2939778} {"why should i trust
  you?" explaining the predictions of any classifier}.
\newblock In \emph{Proceedings of the ACM SIGKDD International Conference on
  Knowledge Discovery and Data Mining}, volume 13-17-August-2016.

\bibitem[{Riquelme et~al.(2018)Riquelme, Tucker, and Snoek}]{Riquelme2018}
Carlos Riquelme, George Tucker, and Jasper Snoek. 2018.
\newblock \href {https://openreview.net/forum?id=SyYe6k-CW} {{Deep Bayesian
  bandits showdown: An empirical comparison of Bayesian deep networks for
  Thompson sampling}}.
\newblock In \emph{Proceedings of the 6th International Conference on Learning
  Representations, {ICLR} 2018, Vancouver, BC, Canada, April 30 - May 3, 2018,
  Conference Track Proceedings}. OpenReview.net.

\bibitem[{Russell(2019)}]{russell2019human}
S.~Russell. 2019.
\newblock \href {https://books.google.co.uk/books?id=M1eFDwAAQBAJ} {\emph{Human
  Compatible: Artificial Intelligence and the Problem of Control}}.
\newblock Penguin Publishing Group.

\bibitem[{Saqaeeyan et~al.(2020)Saqaeeyan, Javadi, and
  Amirkhani}]{Saqaeeyan2020}
Sasan Saqaeeyan, Hamid Haj~Seyyed Javadi, and Hossein Amirkhani. 2020.
\newblock \href {https://doi.org/10.3837/TIIS.2020.04.021} {{Anomaly detection
  in smart homes using Bayesian networks}}.
\newblock \emph{KSII Transactions on Internet and Information Systems}, 14(4).

\bibitem[{Scalise et~al.(2017)Scalise, Rosenthal, and Srinivasa}]{Scalise2017}
Rosario Scalise, Stephanie Rosenthal, and Siddhartha Srinivasa. 2017.
\newblock \href {https://doi.org/10.1145/3029798.3034809} {Natural language
  explanations in human-collaborative systems}.
\newblock In \emph{Proceedings of the Companion of the 2017 ACM/IEEE
  International Conference on Human-Robot Interaction}, HRI '17, page
  377–378, New York, NY, USA. Association for Computing Machinery.

\bibitem[{Schluter(2017)}]{Schluter2017}
Natalie Schluter. 2017.
\newblock \href {https://www.aclweb.org/anthology/E17-2007} {The limits of
  automatic summarisation according to {ROUGE}}.
\newblock In \emph{Proceedings of the 15th Conference of the {E}uropean Chapter
  of the Association for Computational Linguistics: Volume 2, Short Papers},
  pages 41--45, Valencia, Spain. Association for Computational Linguistics.

\bibitem[{Sellam et~al.(2020)Sellam, Das, and Parikh}]{sellam-etal-2020-bleurt}
Thibault Sellam, Dipanjan Das, and Ankur Parikh. 2020.
\newblock \href {https://doi.org/10.18653/v1/2020.acl-main.704} {{BLEURT}:
  Learning robust metrics for text generation}.
\newblock In \emph{Proceedings of the 58th Annual Meeting of the Association
  for Computational Linguistics}, pages 7881--7892, Online. Association for
  Computational Linguistics.

\bibitem[{Smith(2008)}]{Smith2008}
William Smith. 2008.
\newblock \href {https://eric.ed.gov/?id=ED501717} {{Does Gender Influence
  Online Survey Participation? A Record-Linkage Analysis of University Faculty
  Online Survey Response Behavior.}}
\newblock Accessed on 2021-01-25.

\bibitem[{Sokol and Flach(2019)}]{Sokol2019}
Kacper Sokol and Peter~A. Flach. 2019.
\newblock \href {http://ceur-ws.org/Vol-2301/paper\_20.pdf} {Counterfactual
  {E}xplanations of {M}achine {L}earning {P}redictions: {O}pportunities and
  {C}hallenges for {AI} {S}afety}.
\newblock In \emph{Workshop on Artificial Intelligence Safety 2019 co-located
  with the Thirty-Third {AAAI} Conference on Artificial Intelligence 2019
  (AAAI-19), Honolulu, Hawaii, January 27, 2019}, volume 2301 of \emph{{CEUR}
  Workshop Proceedings}. CEUR-WS.org.

\bibitem[{Tashman et~al.(2020)Tashman, Gorder, Parthasarathy, Nasr-Azadani, and
  Webre}]{Tashman2020}
Zaid Tashman, Christoph Gorder, Sonali Parthasarathy, Mohamad~M. Nasr-Azadani,
  and Rachel Webre. 2020.
\newblock \href {https://doi.org/10.3390/su12072897} {{Anomaly detection system
  for water networks in northern ethiopia using bayesian inference}}.
\newblock \emph{Sustainability (Switzerland)}, 12(7).

\bibitem[{Tourigny and Capus(1998)}]{Tourigny1998}
Nicole Tourigny and Laurence Capus. 1998.
\newblock \href {https://doi.org/10.1076/call.11.5.475.5666} {{Learning
  summarization by using similarities}}.
\newblock \emph{International Journal of Phytoremediation}, 21.

\bibitem[{White and d'Avila Garcez(2019)}]{DBLP:journals/corr/abs-1908-03020}
Adam White and Artur~S. d'Avila Garcez. 2019.
\newblock \href {http://arxiv.org/abs/1908.03020} {Measurable counterfactual
  local explanations for any classifier}.
\newblock \emph{CoRR}, abs/1908.03020.

\bibitem[{Winatmoko and Khodra(2013)}]{Winatmoko2013}
Yosef~Ardhito Winatmoko and Masayu~Leylia Khodra. 2013.
\newblock \href {https://doi.org/10.1016/j.protcy.2013.12.290} {{Automatic
  Summarization of Tweets in Providing Indonesian Trending Topic Explanation}}.
\newblock \emph{Procedia Technology}, 11.

\bibitem[{Xu et~al.(2017)Xu, Xing, Xia, and Lo}]{Xu2017}
Bowen Xu, Zhenchang Xing, Xin Xia, and David Lo. 2017.
\newblock \href {https://doi.org/10.1109/ASE.2017.8115681} {{AnswerBot:
  Automated generation of answer summary to developers' technical questions}}.
\newblock In \emph{Proceedings of the 32nd IEEE/ACM International Conference on
  Automated Software Engineering (ASE 2017)}.

\bibitem[{Xu et~al.(2020)Xu, Du{\v{s}}ek, Li, Rieser, and Konstas}]{Xu2020}
Xinnuo Xu, Ond{\v{r}}ej Du{\v{s}}ek, Jingyi Li, Verena Rieser, and Ioannis
  Konstas. 2020.
\newblock \href {https://doi.org/10.18653/v1/2020.acl-main.455} {Fact-based
  content weighting for evaluating abstractive summarisation}.
\newblock In \emph{Proceedings of the 58th Annual Meeting of the Association
  for Computational Linguistics}, pages 5071--5081, Online. Association for
  Computational Linguistics.

\bibitem[{Yuan et~al.(2011)Yuan, Lim, and Lu}]{Yuan:2011:MRE:2208436.2208445}
Changhe Yuan, Heejin Lim, and Tsai-Ching Lu. 2011.
\newblock \href {http://dl.acm.org/citation.cfm?id=2208436.2208445} {Most
  relevant explanation in bayesian networks}.
\newblock \emph{The Journal of Artificial Intelligence Research},
  42(1):309--352.

\bibitem[{Zemla et~al.(2017)Zemla, Sloman, Bechlivanidis, and
  Lagnado}]{Zemla2017}
Jeffrey~C. Zemla, Steven Sloman, Christos Bechlivanidis, and David~A. Lagnado.
  2017.
\newblock \href {https://doi.org/10.3758/s13423-017-1258-z} {Evaluating
  everyday explanations}.
\newblock \emph{Psychonomic Bulletin {\&} Review}, 24(5):1488--1500.

\bibitem[{Zhang et~al.(2020)Zhang, Kishore, Wu, Weinberger, and
  Artzi}]{bert-score}
Tianyi Zhang, Varsha Kishore, Felix Wu, Kilian~Q. Weinberger, and Yoav Artzi.
  2020.
\newblock \href {https://openreview.net/forum?id=SkeHuCVFDr} {{BERTS}core:
  {E}valuating {T}ext {G}eneration with {BERT}}.
\newblock In \emph{International Conference on Learning Representations}.

\end{thebibliography}
\bibliographystyle{acl_natbib}

\end{document}